\documentclass{article}

\PassOptionsToPackage{numbers, compress}{natbib}

\usepackage[final]{tccml_neurips_2020}

\usepackage[utf8]{inputenc} 
\usepackage[T1]{fontenc}    
\usepackage{hyperref}       
\usepackage{url}            
\usepackage{booktabs}       
\usepackage{amsfonts}       
\usepackage{nicefrac}       
\usepackage{microtype}      

\title{The Human Effect Requires Affect: Addressing Social-Psychological Factors of Climate Change with Machine Learning}

\author{%
  Kyle Tilbury \\
  Cheriton School of Computer Science\\
  University of Waterloo\\
  Waterloo, Ontario, Canada\\
  \texttt{ktilbury@uwaterloo.ca} \\
  \And
  Jesse Hoey \\
  Cheriton School of Computer Science\\
  University of Waterloo\\
  Waterloo, Ontario, Canada\\
  \texttt{jhoey@cs.uwaterloo.ca} \\
}

\begin{document}
\maketitle


\section{Introduction}

Climate change is a global collective action problem where human populations and the Earth’s climate are engaged in a two-way feedback loop. This feedback loop is characterized by how climate change impacts human decisions and behaviours and how in-turn human decisions and behaviours impact climate change. Not accounting for human effects when designing approaches that grapple with climate change is like not considering traffic when designing a bridge~\cite{palmer2014model}.
It has been established that, just as there are human effects aggravating climate change, there are human actions that can be taken to lower the impacts of climate change on the Earth and individuals.

Researchers have proposed applying machine learning (ML) to aid in mitigating the human effects in climate change. Proposed applications of machine learning include identifying to consumers which of their behaviours result in the highest emissions and strategies to reduce them, suggesting the adoption of new technologies to those individuals  most amenable to them, and targeting informational interventions to high emission households~\cite{rolnick2019tackling}. 
The success of these machine learning efforts, however, will largely depend on their capacity to effectively impact human climate change perceptions and behaviour.

The groundwork for understanding the factors that dictate people's climate change perceptions and behaviours has been established by research in sociology and psychology. A key driver among these human factors, currently overlooked in most ML for climate change research, is affect. Affect typically refers to the underlying experience of feelings or emotions and can be thought of as an evaluative heuristic that influences information processing~\cite{zajonc1980feeling}. Attention-catching and emotionally engaging informational interventions can be impactful in generating individual or collective action in response to climate change~\cite{weber2010shapes}. Statistical information about climate change, presented by itself, means very little to people as it requires analytic effortful processing~\cite{weber2010shapes}. However, information presented about climate change in an affective manner is impactful in improving people's engagement with climate change, as affective processing of information is fast, somewhat automatic, and motivates action~\cite{van2015improving,weber2010shapes}. It stands to reason that ML approaches for the climate struggle will benefit from interventions that engage with humans in a similarly affective manner.

We propose that leveraging social-psychological factors, specifically affect, will play a critical role in allowing artificially intelligent agents to interact with humans in a way that maximizes climate change mitigation efforts. 
We expect that a ML based approach that apprises an individual of some information, such as a strategy to reduce their household emissions, while taking into account social-psychological factors of that individual, like affect, will be more likely to facilitate a behavioural change than many existing approaches. 
Rather than presenting statistical summary data and generic strategies to reduce the individual's carbon footprint, the application can encourage or “nudge”~\cite{Nudge2008} them to engage in a mitigation strategy based on that individual's learned affective identity and other relevant social-psychological factors. 
For example, a middle-class ``manly'' electrician might install solar panels, but would never give up his big truck, as he considers it part of his identity. 
So, interventions to reduce emissions by encouraging a more fuel-efficient vehicle could be futile and would be better targeted elsewhere. 
We envision that using affective machine learning will increase the efficacy of informational interventions and enhance the adoption of mitigative strategies at the individual and group level.

\section{Human Social-Psychological Factors of Climate Change}
Social-psychological factors that play a role in climate change have been identified in models of human perceptions and willingness to engage in mitigation behaviours~\cite{van2015social,xie2019predicting}. These social-psychological factors are broken down into four categories: (1) socio-demographics including age, gender, and political affiliation; (2) cognitive factors including knowledge of climate change causes, impacts, responses, and response efficacy; (3) experiential processes including personal experience and affect; and (4) socio-cultural influences including social norms and cultural values. These social-psychological factors identified predict perceptions and behaviour willingness but are not direct predictors of actual behaviour. However, perceptions and behaviour willingness are direct predictors of actual behaviour~\cite{ajzen1991theory, o1999risk}. These models are largely derived using data from Western industrialized democracies and social-psychological factors at play in people's perceptions and intentions towards climate change may differ in other populations. Of the identified social-psychological factors relevant to climate change, affect was found to be the single largest predictor~\cite{van2015social,xie2019predicting}. This justifies researcher's recommendations that policy-makers emphasize personal and affective climate change interventions~\cite{van2015social,van2015improving,weber2010shapes,xie2019predicting} and justifies our call for affect to be incorporated in applications of ML that deal with human effects in climate change.

\section{Incorporating Affect in Machine Learning for Climate Change}

To endow an artificial agent with the ability to have affectively intelligent interactions in the climate change setting, we propose utilizing an existing approach for affective AI called Bayesian affect control theory (BayesACT)~\cite{hoey2016affect,schroder2016modeling}. BayesACT is based on a comprehensive social psychological theory of human social interaction sociological known as affect control theory~\cite{heise2007expressive}. BayesACT agents are capable of learning affective identities of their interactants, predicting how the affective state of an interaction will progress, and how to select a strategy of action in order to maximize the expected value of an outcome while incorporating affective alignment with the human~\cite{hoey2016affect}. We outline two approaches for using BayesACT to gain insights into how affective AI could aid in climate change mitigation.

\subsection{Affective Agent-based Modelling for Climate Change}
In order to understand how individuals and groups may act in response to different interactions with affectively intelligent agents we propose to use agent-based modelling~\cite{bonabeau2002agent}.
Agent-based models simulate the actions and interactions of agents in a human based system to understand dynamics of that system. 
As the models reflect the real world, we can use them to learn behaviours and actions. 
With respect to climate change, agent-based modelling has previously been used to study dynamics of social-psychological factors including social norms~\cite{bury2019charting,schluter2016robustness} and political and economic beliefs~\cite{geisendorf2018evolutionary}. We propose utilizing systems and models in which different mitigation strategies and behaviours are modelled in relation to expected utilities, social preferences, and climate policy structures, but with the addition of affective factors. BayesACT is well suited for this as it can model economic and monetary factors in addition to social and emotional ones. It can explicitly represent the structure of social interactions based on identity, coupled with the factual meanings embedded in information garnered from climate data and policy structures. 
In other contexts, BayesACT has been shown to model complex large-scale human interactions and status processes through structures of deference~\cite{freeland2018structure}. The implications of this are predictions about human action. For example, if your neighbour is someone you respect because of their occupational status, you may be more likely to imitate them. So, we expect BayesACT to be able to model similarly complex interactions in a climate change setting. This agent-based modelling approach could elucidate how intelligently delivered affective interventions can change mitigation strategies at the individual and group level. That is, we will be able to observe how the use of affect can influence the adoption of climate change mitigation efforts at scale. 

\subsection{A Climate Change Social Dilemma with People and Affective Agents}
To gain real-world insights on affectively intelligent agents for climate change mitigation, we propose a simple small-scale experiment involving humans. Climate change can be considered as a collective-risk social dilemma and can be simulated as a game with a group of people~\cite{milinski2008collective}. The game consists of some number of people who can contribute ``money'' over successive rounds of the game to a mitigative fund. If a certain threshold is met at the end of the game, players get to keep their remaining money that they have not contributed to the fund. If the threshold of mitigation is not met, all players lose their remaining money with a certain probability. This probability is an analog to the perceived risk of loss associated with climate change. Human groups manage to reach the target sum and avoid simulated climate catastrophe only when they know that the risk of loss is high.

We propose an adaptation of this game where each human interacts with an assistant agent to try and avoid this social dilemma. In addition to contributing each round, each human has an assistant that attempts to deliver informational interventions to guide mitigative behaviours so that the goal is reached. Our goal would be for agents to aid the humans in reaching the mitigation threshold even when the perceived risk of loss is low. To evaluate the effect of affect we propose using both affective agents and a baseline agent that serves as a control in the social game experiment. Throughout the rounds of the game, the affective agent could learn the affective identity of the participant and deliver tailored messages aligned with that identity. For example, some individuals may be more receptive to messages that are encouraging or others more responsive to critical messages. This would be contrasted with generic messages delivered by the baseline agent that does not incorporate affect. The results from this experiment will be a proof of concept of affective agents aiding humans in avoiding toy collective-risk social dilemmas. Small scale experiments such as this can be useful in determining dynamics that can have implications at a global scale~\cite{milinski2008collective}.

\subsection{Complications}

Despite the potential for our proposal to aid in climate change mitigation, there is the possibility of unintended side-effects. One such side-effect is the single-action bias~\cite{weber2010shapes}. This is a phenomena where an individual takes an action to mitigate a risk that they worry about, but then they have a tendency to not take any further action as the first action suffices in reducing their worry about the risk. So, inducing an individual to take a mitigative action towards climate change has the potential to make them less likely to take further mitigation actions. Another possible downside is the finite pool of worry effect~\cite{weber2010shapes}. This is the phenomena where when an individual's worry increases about one type of risk, concern for other risks decrease. So, increasing concerns about climate change may decrease concerns for other relevant risks. There is also potential for negative uses of affectively intelligent ML methods in general. If affectively intelligent AI is capable of influencing perceptions and behaviours for the benefit of climate change mitigation, it seems just as likely that the technology could be used to act against climate change mitigation or in any number of nefarious ways. The real-world use of these methods should be predicated on a number of relevant questions relating to privacy, trust, and ethical use.

\section{Conclusion}

Human's perceptions and their willingness to engage in mitigative behaviours impact Earth's climate. 
We propose that machine learning based approaches that contend with humans and climate change must incorporate affect. We outline an investigation into how affect could be utilized to enhance machine learning for climate change. Behavioural and informational interventions can be a powerful tool in helping humans adopt mitigative behaviours, especially when they are affectively engaging. Utilizing ML to automatically construct and deliver affective interventions could help mitigative behaviours become widely adopted. Long term, we envision that socially and psychologically aware AI can aid climate change mitigation and adaptation efforts at the regional, national, and international levels. For example, the AI could automatically monitor extreme weather events, advise local farmers on medium-term strategies in ways that best suit each farmer's emotional and social factors for preparation, give them better estimates of anticipated changes, and catalyze decisions to act cooperatively with other entities in the food supply chain to temper losses. Understanding how AI can coordinate, communicate, and cooperate with humans, while being aligned in a social-psychological manner, will be beneficial in tackling climate change at a global scale.

\bibliographystyle{plainnat}
\bibliography{biblio}

\begin{thebibliography}{19}
\providecommand{\natexlab}[1]{#1}
\providecommand{\url}[1]{\texttt{#1}}
\expandafter\ifx\csname urlstyle\endcsname\relax
  \providecommand{\doi}[1]{doi: #1}\else
  \providecommand{\doi}{doi: \begingroup \urlstyle{rm}\Url}\fi

\bibitem[Ajzen(1991)]{ajzen1991theory}
Icek Ajzen.
\newblock The theory of planned behavior.
\newblock \emph{Organizational Behavior and Human Decision Processes},
  50\penalty0 (2):\penalty0 179--211, 1991.

\bibitem[Bonabeau(2002)]{bonabeau2002agent}
Eric Bonabeau.
\newblock Agent-based modeling: Methods and techniques for simulating human
  systems.
\newblock \emph{Proceedings of the National Academy of Sciences}, 99\penalty0
  (suppl 3):\penalty0 7280--7287, 2002.

\bibitem[Bury et~al.(2019)Bury, Bauch, and Anand]{bury2019charting}
Thomas~M Bury, Chris~T Bauch, and Madhur Anand.
\newblock Charting pathways to climate change mitigation in a coupled
  socio-climate model.
\newblock \emph{PLoS computational biology}, 15\penalty0 (6), 2019.

\bibitem[Freeland and Hoey(2018)]{freeland2018structure}
Robert~E Freeland and Jesse Hoey.
\newblock The structure of deference: Modeling occupational status using affect
  control theory.
\newblock \emph{American Sociological Review}, 83\penalty0 (2):\penalty0
  243--277, 2018.

\bibitem[Geisendorf(2018)]{geisendorf2018evolutionary}
Sylvie Geisendorf.
\newblock Evolutionary climate-change modelling: A multi-agent climate-economic
  model.
\newblock \emph{Computational Economics}, 52\penalty0 (3):\penalty0 921--951,
  2018.

\bibitem[Heise(2007)]{heise2007expressive}
David~R Heise.
\newblock \emph{Expressive order: Confirming sentiments in social actions}.
\newblock Springer Science \& Business Media, 2007.

\bibitem[Hoey et~al.(2016)Hoey, Schr{\"o}der, and Alhothali]{hoey2016affect}
Jesse Hoey, Tobias Schr{\"o}der, and Areej Alhothali.
\newblock Affect control processes: Intelligent affective interaction using a
  partially observable markov decision process.
\newblock \emph{Artificial Intelligence}, 230:\penalty0 134--172, 2016.

\bibitem[Milinski et~al.(2008)Milinski, Sommerfeld, Krambeck, Reed, and
  Marotzke]{milinski2008collective}
Manfred Milinski, Ralf~D Sommerfeld, Hans-J{\"u}rgen Krambeck, Floyd~A Reed,
  and Jochem Marotzke.
\newblock The collective-risk social dilemma and the prevention of simulated
  dangerous climate change.
\newblock \emph{Proceedings of the National Academy of Sciences}, 105\penalty0
  (7):\penalty0 2291--2294, 2008.

\bibitem[O'Connor et~al.(1999)O'Connor, Bard, and Fisher]{o1999risk}
Robert~E O'Connor, Richard~J Bard, and Ann Fisher.
\newblock Risk perceptions, general environmental beliefs, and willingness to
  address climate change.
\newblock \emph{Risk Analysis}, 19\penalty0 (3):\penalty0 461--471, 1999.

\bibitem[Palmer and Smith(2014)]{palmer2014model}
Paul~I Palmer and Matthew~J Smith.
\newblock Model human adaptation to climate change.
\newblock \emph{Nature}, 512\penalty0 (7515):\penalty0 365, 2014.

\bibitem[Rolnick et~al.(2019)Rolnick, Donti, Kaack, Kochanski, Lacoste,
  Sankaran, Ross, Milojevic-Dupont, Jaques, Waldman-Brown,
  et~al.]{rolnick2019tackling}
David Rolnick, Priya~L Donti, Lynn~H Kaack, Kelly Kochanski, Alexandre Lacoste,
  Kris Sankaran, Andrew~Slavin Ross, Nikola Milojevic-Dupont, Natasha Jaques,
  Anna Waldman-Brown, et~al.
\newblock Tackling climate change with machine learning.
\newblock \emph{arXiv preprint arXiv:1906.05433}, 2019.

\bibitem[Schl{\"u}ter et~al.(2016)Schl{\"u}ter, Tavoni, and
  Levin]{schluter2016robustness}
Maja Schl{\"u}ter, Alessandro Tavoni, and Simon Levin.
\newblock Robustness of norm-driven cooperation in the commons.
\newblock \emph{Proceedings of the Royal Society B: Biological Sciences},
  283\penalty0 (1822):\penalty0 20152431, 2016.

\bibitem[Schr{\"o}der et~al.(2016)Schr{\"o}der, Hoey, and
  Rogers]{schroder2016modeling}
Tobias Schr{\"o}der, Jesse Hoey, and Kimberly~B Rogers.
\newblock Modeling dynamic identities and uncertainty in social interactions:
  Bayesian affect control theory.
\newblock \emph{American Sociological Review}, 81\penalty0 (4):\penalty0
  828--855, 2016.

\bibitem[Thaler and Sunstein(2008)]{Nudge2008}
Richard~H. Thaler and Cass~R. Sunstein.
\newblock \emph{Nudge: Improving Decisions about Health, Wealth, and
  Happiness}.
\newblock Yale University Press, April 2008.

\bibitem[Van~der Linden(2015)]{van2015social}
Sander Van~der Linden.
\newblock The social-psychological determinants of climate change risk
  perceptions: Towards a comprehensive model.
\newblock \emph{Journal of Environmental Psychology}, 41:\penalty0 112--124,
  2015.

\bibitem[Van~der Linden et~al.(2015)Van~der Linden, Maibach, and
  Leiserowitz]{van2015improving}
Sander Van~der Linden, Edward Maibach, and Anthony Leiserowitz.
\newblock Improving public engagement with climate change: Five “best
  practice” insights from psychological science.
\newblock \emph{Perspectives on Psychological Science}, 10\penalty0
  (6):\penalty0 758--763, 2015.

\bibitem[Weber(2010)]{weber2010shapes}
Elke~U Weber.
\newblock What shapes perceptions of climate change?
\newblock \emph{Wiley Interdisciplinary Reviews: Climate Change}, 1\penalty0
  (3):\penalty0 332--342, 2010.

\bibitem[Xie et~al.(2019)Xie, Brewer, Hayes, McDonald, and
  Newell]{xie2019predicting}
Belinda Xie, Marilynn~B Brewer, Brett~K Hayes, Rachel~I McDonald, and Ben~R
  Newell.
\newblock Predicting climate change risk perception and willingness to act.
\newblock \emph{Journal of Environmental Psychology}, 65:\penalty0 101331,
  2019.

\bibitem[Zajonc(1980)]{zajonc1980feeling}
Robert~B Zajonc.
\newblock Feeling and thinking: Preferences need no inferences.
\newblock \emph{American Psychologist}, 35\penalty0 (2):\penalty0 151, 1980.

\end{thebibliography}

\end{document}